# Introducing New AdaBoost Features for Real-Time Vehicle Detection

Bogdan Stanciulescu (1), Amaury Breheret (2), Fabien Moutarde (3)
1 : Robotics Laboratory – Ecole des Mines de Paris, France, 60 Boulevard Saint-Michel, 75272 Paris, France, bogdan@ensmp.fr
2 : Robotics Laboratory – Ecole des Mines de Paris, breheret@ensmp.fr, 3 : Robotics Laboratory – Ecole des Mines de Paris, moutarde@ensmp.fr

**Introduction**

This paper shows how to improve the real-time object detection in complex robotics applications, by exploring new visual features as AdaBoost weak classifiers. Previous work on AdaBoost resulted in two generic weak-classifier feature categories, the Haar-like feature and the control-points feature. Although they are well suited for un-structured object detection, such as faces and pedestrians, their performances are limited when detecting highly structured objects. This paper extends the existing visual features through inserting topological properties, such as symmetry, contiguity, vertical and horizontal limits (edges), corners, etc., and explores new categories of features.

The case study (benchmark) is the generic vehicle detection in a dynamic and changing environment. Vehicles are more and more considered as being advanced robots, by both their behavior and their shapes. Contests such as DARPA Challenge are pushing vehicle technology towards autonomous robotics. Vehicles are equipped by exteroceptive and proprioceptive sensors; their information is used either for navigation purposes or for obstacle detection and avoidance. One of the most important exteroceptive sensors is the mono-camera vision sensor.

Complete tests are performed on ground-truth sequences, showing the real-time detection performances brought by each kind of the visual features.

**AdaBoost basics**

The AdaBoost algorithm was introduced in 1995 by Y. Freund and R. Shapire [1]. The algorithm description is based on [2], of the above authors. The algorithm takes as input a training set made of $(x_1,y_1),…,(x_m,y_m)$ examples. The $x_i$ are examples to be classified, in a $X$-domain or space, whereas $y_i$ are classes. For simplicity, we assume dealing with two classes of objects, $Y = \{-1, 1\}$. The main idea of the algorithm is to construct a probability distribution or a set of weights over the training set, using a repetitive procedure called *weak learner* in series of steps $t = 1,…T$. The weight of this distribution on the $i$th example is $D_t(i)$. The example weights are initially set equal, but at each step they are modified so the incorrectly classified examples have their weights increased. This forces the weak learner to focus on the hard examples, at the next step. The weak learner's job is to find a *weak classifier* $h_t = X \to \{-1,+1\}$ that minimize the classification error according to the distribution $D_t$, over the training example set.

$$\varepsilon_t = \sum_{i=1}^{I} weight(h_t(i)) \begin{Bmatrix} if(h_t(i) \neq y_i) \\ else(0) \end{Bmatrix}$$

Every weak classifier will have a weight that depends of the classification error. Lesser the classification error $\varepsilon_t$, bigger is the weight $\alpha_t$ accorded to that weak classifier in the final combination will be.

$$\alpha_t = \ln\left(\frac{1-\varepsilon_t}{\varepsilon_t}\right)$$

At the end of the algorithm every weak classifier is weighted by the computed $\alpha_t$ weight, during the learning stage.

Starting from now, every new example $x$ is classified according to a weighted voting rule, usually called *strong classifier*:

$$H(x) = sign\left(\sum_{t=1}^{T} \alpha_t h_t(x)\right)$$

We can notice that AdaBoost is constructing a strong classifier as a sum of basic functions, which are the weak classifiers. By adapting the weight distribution at each step $t$ in order to enforce the classification of the remaining wrong-classified examples, AdaBoost is producing orthogonal basic rules, or weak classifiers, as depicted in the figure bellow [3].

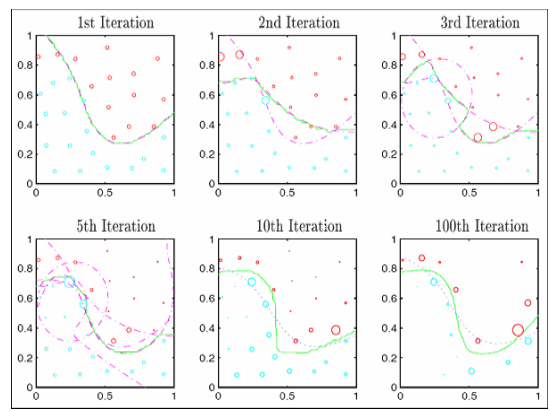

The pseudo-code of the AdaBoost training algorithm is presented in the figure here-after.

---

**Inputs** : A set of N examples $X = \{(x_1, y_1), ..., (x_N, y_N)\}$ with $y_i \in \{-1, 1\}$, number of rounds T, and a genetic algorithm weak learner.
**Outputs** : A strong classifier H(x).
1. Initialise $D_1^i = \frac{1}{N}$ for i = 1, ..., N
2. **for** t = 1, 2, ..., T **do**
3.     Train the genetic weak learner using distribution $D_t$ and obtain weak classifier $h_t$
4.     Compute classification error $\varepsilon_t$ of $h_t$ :
$$\varepsilon_t = \sum i : h_t(x_i) \neq y_i D_t^i$$
5.     **if** $\varepsilon_t = 0$ **or** $\varepsilon_t > \frac{1}{2}$ **then**
6.         Exit the loop and put T = t − 1.
7.     **else**
8.         define $\beta_t = \frac{\varepsilon_t}{1-\varepsilon_t}$
9.     Update the weights distribution
$$D_t : D_{t+1}^i \leftarrow \frac{D_t^i}{Z_t} x \begin{cases} \beta_t, if(h_t(x_i) = y_i) \\ 0, else \end{cases} \text{ with } Z_t$$
a normalization term for $D_{t+1}$ distribution.
10.     **endif**
11. **endfor**
12. Output $H(x) = \sum_{t=1}^{T} \alpha_t H_{t(x)}$ with $\alpha_t = \log(\frac{1}{\beta_t})$

---

**AdaBoost features for object detection**
Several machine-vision algorithms were developed during the last decade, but among them only few are able to provide real-time compatibility. The boosting algorithms were successfully extended to machine-vision by Viola & Jones [4][5]. They introduced the Haar-like features as AdaBoost week-classifiers, for face and pedestrian detection. The resulting detector is running at 4 images per second, using motion information.

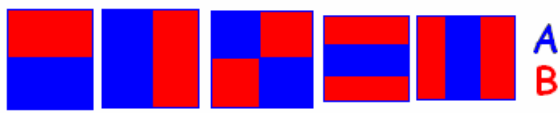

*Viola & Jones Haar-like features for pedestrian and face detection*

The weak classifier introduced by these features computes the absolute difference between the sum of pixel values in red and blue areas, with the respect of the following rule:
**if** $|Area(A) - Area(B)| > Threshold$ **then True else False**

In order to ensure an ilumination independent characteristic of the algorithm, one must normalise the areas by using pixel variance deduced from the integral of square of image, before applying these features.

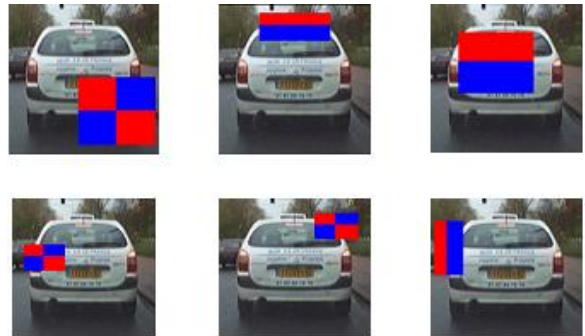

*AdaBoost trained Viola-Jones features for vehicle detection. This result was obtained from 500 positive and 1000 negative examples set training*

Abramson & Steux [6][7] proposed a faster method based on a different type of feature, the control-points. This feature operates directly at pixel level and is illumination-independent. The pre-processing time needed for an integral image computation and a histogram-equalization is saved, providing very good real-time performances. Arbitrary points are divided in two sub-classes, one called the positive set and the second called the negative set.

Examples to be classified as positive, respect the following rule:

$$\min(P_i^+) - \max(P_i^-) > V \text{ OR}$$
$$\min(P_i^-) - \max(P_i^+) > V \text{ FOR } i = 1,2,...N$$

*V* is the separation threshold between the two point classes, $P^+i$ a point from the positive class and $P^-i$ a point from the negative class, and *N* the number of points in each class.

In a linear representation of the pixel values, a positive-classified example is the one having the two classes separated by a threshold *V*. A negative example is the one that does not respect this characteristic: values of the control-points are not separated in two distinct classes.

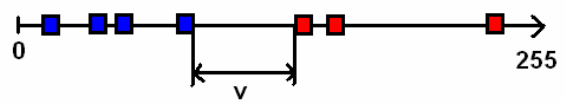

*Positive-classified example with respect to the*

*threshold V.*

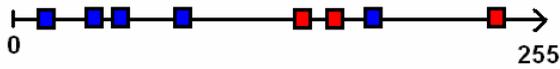

*Negative-classified example.*

One can see in the figure bellow how control-points are acting in a car-detection example.

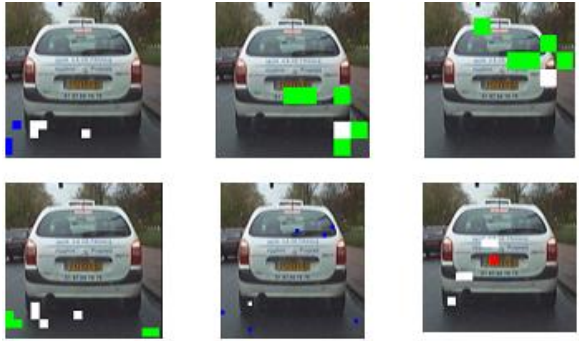

*AdaBoost trained Control-Points features for vehicle detection. This result was obtained from 500 positive and 1000 negative examples set training.*

**Introducing new features**

While on pedestrian and face detection the control-points are slightly better than the Haar-like features, on vehicle-detection applications they are surpassing them significantly. Both features are simple to implement and they are bringing satisfying detection results, but they are not exploiting the vehicle structure characteristics, such as symmetry, vertical and horizontal contours, corners, etc.

Another inconvenient of these features is their lack of visual meaning. As one can see in the figures above, the features resulted from the AdaBoost training are difficult, if not impossible, to interpret. The relation between the feature performance and their signification is not always easy to seize. This paper challenge is to improve, not only the detection performance, but also the direct relation between the best features and their visual meaning.

Taking the above considerations as the start-point of our research, we have proposed and developed two types of weak features: symmetrical Viola-Jones and N-connexity control-points. Tests have been conducted on a ground-truth sequence proving their efficiency.

**Symmetrical Viola-Jones feature**

Related to Viola-Jones, the symmetrical Viola-Jones feature exploits the properties of vertical symmetry and the vertical and horizontal borders or edges, the vehicles have.

Symmetrical Viola-Jones is much more complex than the original one. It is composed of three different Viola-Jones original features, one in the middle of the image and another two symmetrically positioned at the right and the left of the first. We denote them by $Z_1$, the left area, $Z_2$, the right area and by $Z_3$ the middle area. By the index A and B we denote the two areas composing each individual Viola-Jones feature. A positive-classified example should respect five conditions. By defining,

$$Diff_1 = \left|Area(Z_1^A) - Area(Z_1^B)\right|$$
$$Diff_2 = \left|Area(Z_2^A) - Area(Z_2^B)\right|$$
$$Diff_3 = \left|Area(Z_3^A) - Area(Z_3^B)\right|$$

we have the following conditions:

Condition 1 :
$$(Diff_1 > Threshold(Z_1))$$
Condition 2 :
$$(Diff_2 > Threshold(Z_2))$$
Condition 3 :
$$(Diff_3 > Threshold(Z_3))$$
Condition 4 :
$$\left|Diff_1 - Diff_2\right| < ThresholdDiff1$$
Condition 5 :
$$\left|\left(\left|Diff_1 - Diff_2\right|\right) - Diff_3\right| > ThresholdDiff2$$

One can see that this type of classifier uses five different thresholds. The first three of them are the same classical thresholds as in original Viola-Jones, while the next two are thresholds related to the symmetry and the anti-symmetry conditions. The fourth threshold forces the symmetry between the left and right Viola-Jones features. They must be close enough in order to be considered symmetrical. The fifth threshold imposes a minimal difference of the middle Viola-Jones feature, compared with the extremities. During the training, the weak learner, based on a genetic algorithm and further described in [8], is constructing various Symmetrical Viola-Jones features, which are injected in the AdaBoost loop.

As one can see in the following figure, the weak classifiers based on this new feature manage to find symmetrical edges of the vehicle.

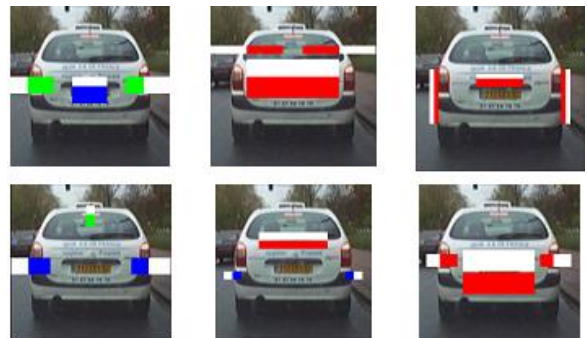

*Symmetrical Viola-Jones features for vehicle detection. This result was obtained from 500 positive and 1000 negative examples set training.*

**N-connexity feature**

This feature is a particular form of the control-points feature. It contains 2 up to 12 points ordered in an eight-connexity form. Vehicles are rigid and contiguous objects therefore conditions of similar nature should be imposed to the classical control-points feature, in order to decrease the size of the search-space. The classical control-points feature with 2 up to 12 points is generating a search-space of $10^{36}$ possible configurations. By imposing the eight-connexity constraint, the search-space size decreases to $3 \times 10^{19}$ possible combinations.

In the figure bellow one can see some examples of the N-connexity feature as resulted from the AdaBoost training process on a vehicle image-base.

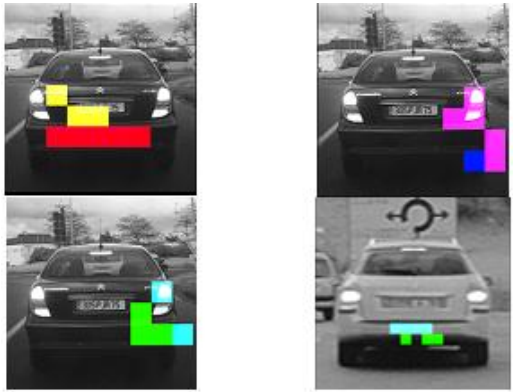

*N-connexity features for vehicle detection.*

**Results and conclusions**

For each type of the proposed features an AdaBoost detector of 500 weak-classifiers was trained. We have tested the detection performances using a ground-truth sequence made from 5000 images, containing vehicles with know positions and sizes.

The results are depicted as both ROC-curve and Precision-Recall characteristics. The ROC (Receiver Operating Characteristics) is the most used characteristic curve in industrial detection applications. It depicts the dependence of the *true positive detection rate* (ordinate axis) with the *false detection rate* (abscissa axis). The Precision-Recall characteristic is mostly used in scientific contests. The *Recall* term denotes the same quantity as the *true positive detection rate*, here on the abscissa axis. The *Precision* term corresponds to the number of true positive detections divided by the total number of positive detections, including the false positive ones. A ground-truth image-sequence was used to evaluate the performance of the each type of feature-based detectors.

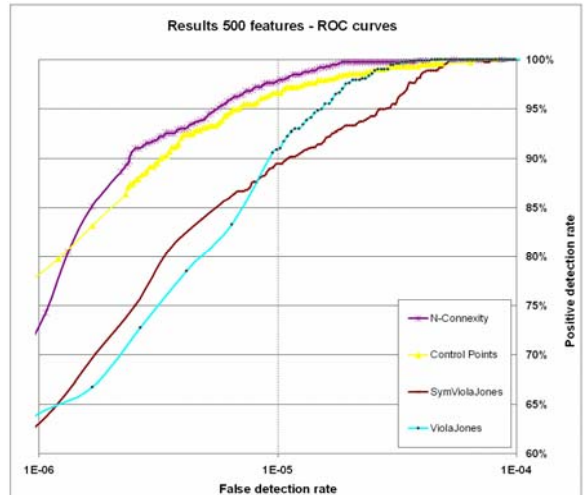

*ROC-curves corresponding to the new features.*

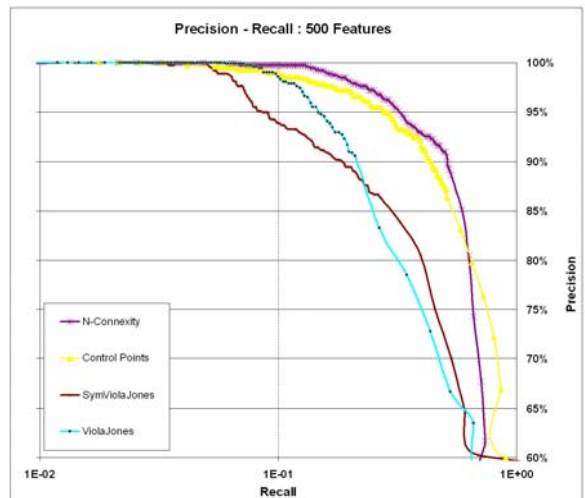

*Precision-recall curves corresponding to the new features.*

As one can see in the above figures the new N-connexity feature generates the best detection results. The Symmetrical Viola-Jones perfoms better than the classical Viola-Jones, but it is still not as good as the classical control-points. We can conclude that for the vehicle-detection problem, the pixel-based features perform globally better that the haar-based features; among them the N-connexity having the best results. However Symmetrical Viola-Jones holds the interesting property to detect object borders. This information could be exploited in order to delimit the regions of interest for other exteroceptive sensors, such as radars or lidars.

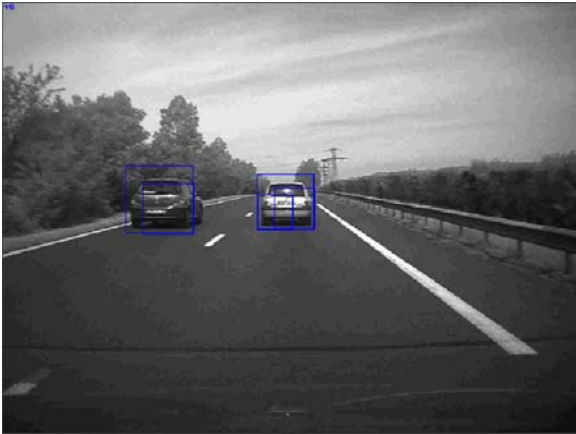

*Vehicle detector using 500 N-connexity AdaBoost features*